\documentclass[10pt,twocolumn,letterpaper]{article}

\usepackage[pagenumbers]{cvpr} 

\definecolor{cvprblue}{rgb}{0.21,0.49,0.74}
\usepackage[pagebackref,breaklinks,colorlinks,allcolors=cvprblue]{hyperref}
\usepackage{multirow,booktabs}
\usepackage{graphicx}
\usepackage{CJKutf8}
\usepackage{capt-of}

\def\paperID{***} 
\def\confName{CVPR}
\def\confYear{2026}

\title{LogoDiffuser: Training-Free Multilingual Logo Generation and Stylization\\via Letter-Aware Attention Control}

\author{
Mingyu Kang$^{1*}$ \quad
Hyein Seo$^{2*}$ \quad
Yuna Jeong$^{1}$ \quad
Junhyeong Park$^{1}$ \quad
Yong Suk Choi$^{2\dagger}$ \\
$^{1}$Department of Artificial Intelligence, Hanyang University \\
$^{2}$Department of Computer Science, Hanyang University \\
{\tt\small 
\{alsrb15788, appleshi, dbsdk, junhyeong820, cys\}@hanyang.ac.kr
} \\
{\small *Equal contribution \quad $\dagger$Corresponding author}
}

\begin{document}

\twocolumn[{
\renewcommand\twocolumn[1][]{#1}
\maketitle

\begin{center}
    \centering
    \vspace*{-.15cm}
    \includegraphics[width=\textwidth]{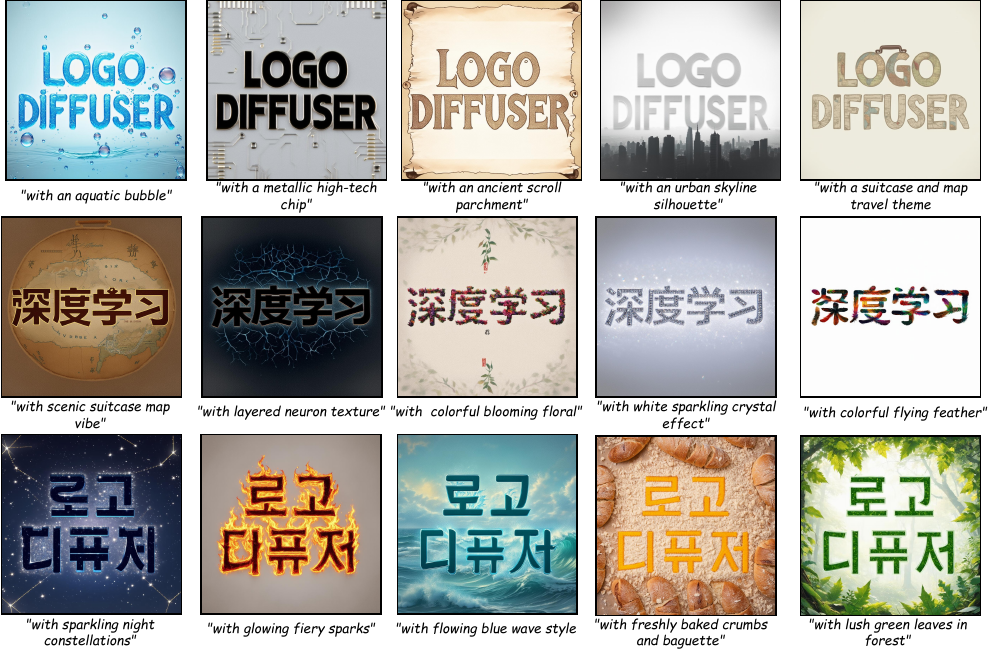}
    \vspace*{-.6cm}
    \captionof{figure}{Our logo generation results on the MM-DiT architecture, showing high-quality outputs across diverse style prompts. The corresponding style information used for image generation is provided within each prompt, which is displayed below each image.
    }
\label{fig:abstract}
\end{center}
}]

\begin{abstract}
Recent advances in text-to-image generation have been remarkable, but generating multilingual design logos that harmoniously integrate visual and textual elements remains a challenging task. Existing methods often distort character geometry when applying creative styles and struggle to support multilingual text generation without additional training.
To address these challenges, we propose LogoDiffuser, a training-free method that synthesizes multilingual logo designs using the multimodal diffusion transformer. Instead of using textual prompts, we input the target characters as images, enabling robust character structure control regardless of language. We first analyze the joint attention mechanism to identify “core tokens”, which are tokens that strongly respond to textual structures. With this observation, our method integrates character structure and visual design by injecting the most informative attention maps.
Furthermore, we perform layer-wise aggregation of attention maps to mitigate attention shifts across layers and obtain consistent core tokens. Extensive experiments and user studies demonstrate that our method achieves state-of-the-art performance in multilingual logo generation.

\end{abstract}    
\section{Introduction}
\label{sec:intro}

Logo images visually represent a brand’s identity and serve as a crucial design element for enhancing recognition of products or services. In the global market, logo designs must achieve both linguistic diversity and visual consistency, motivating the need for technologies that can automatically generate multilingual logo designs integrating text and graphics harmoniously.

Recent advancements in text-to-image generation models \cite{ramesh2021zero, rombach2022high, saharia2022photorealistic, podell2023sdxl} have greatly expanded the boundaries of visual creativity. In particular, the combination of flow models \cite{liu2022flow, lipman2022flow, xu2022poisson} and multimodal diffusion transformers (MM-DiT) \cite{esser2024scaling, peebles2023scalable, chen2024pixart} enables the precise synthesis of complex visual scenes from textual descriptions. However, despite these advancements, the generation of visual text remains an unsolved challenge \cite{ramesh2021zero, saharia2022photorealistic, hu2023tifa, rombach2022high}. In logo design, it is essential to preserve fine-grained textual structures \cite{azadi2018multi, park2021multiple, wang2015deepfont} such as strokes, serifs, and curves, while maintaining stylistic coherence.

Recent studies have attempted to address this issue. Several methods \cite{chen2023textdiffuser, chen2024textdiffuser} employ pre-learned text layout priors to guide the model in generating text within specified regions. While effective in constraining placement, these layout-dependent approaches limit compositional flexibility and may lead to unnatural results when misaligned with actual design principles. Other approaches \cite{yang2023glyphcontrol, tuo2023anytext} render text as glyph images and then insert them into the generated scene. However, these methods often suffer from disrupted visual harmony or distorted character shapes \cite{wu2019editing, roy2020stefann}. Moreover, handling multilingual characters without separate training remains challenging for most existing approaches \cite{azadi2018multi}.

To overcome these limitations, we propose LogoDiffuser, a novel training-free method for generating multilingual logo designs directly within MM-DiT \cite{esser2024scaling}. Instead of using textual prompts alone \cite{nichol2021glide, saharia2022photorealistic}, LogoDiffuser utilizes the target characters as image inputs, allowing precise control over character structures regardless of language. We analyze the importance of different components within the joint self-attention of MM-DiT to identify tokens that strongly respond to character shapes. Specifically, our method measures the variance of token-wise attention scores during character image reconstruction across layers to automatically detect ``core tokens'', which are essential for preserving textual structures. We observe that these core tokens play a pivotal role in balancing structural fidelity and stylistic expression. By selectively injecting their features, LogoDiffuser renders accurate character forms while naturally integrating the desired visual style.

Furthermore, we observe an attention shift in deeper layers, where core tokens gradually distribute attention to non-text regions such as the background. To mitigate this, we introduce a layer-wise attention aggregation strategy that accumulates and averages attention maps across layers, yielding stable and consistent token representations. This approach maintains structural integrity while enabling creative stylistic transformations, as illustrated in Figure \ref{fig:abstract}. Extensive experiments and user studies demonstrate that LogoDiffuser achieves high design quality and text accuracy, highlighting strong potential for multilingual logo generation.

Our main contributions are summarized as follows:
\begin{itemize}
    \item We propose LogoDiffuser, a training-free method for multilingual logo design that treats text as image input, enabling balanced and creative integration between textual and visual elements.
    \item We provide the analysis of MM-DiT’s attention mechanism, finding that certain tokens concentrate around character regions and play a pivotal role in injecting textual structure.
    \item We improve both text accuracy and visual fidelity, achieving precise and visually diverse multilingual logo generation.
\end{itemize}

\begin{figure*}[t]
  \centering
   \includegraphics[width=\linewidth]{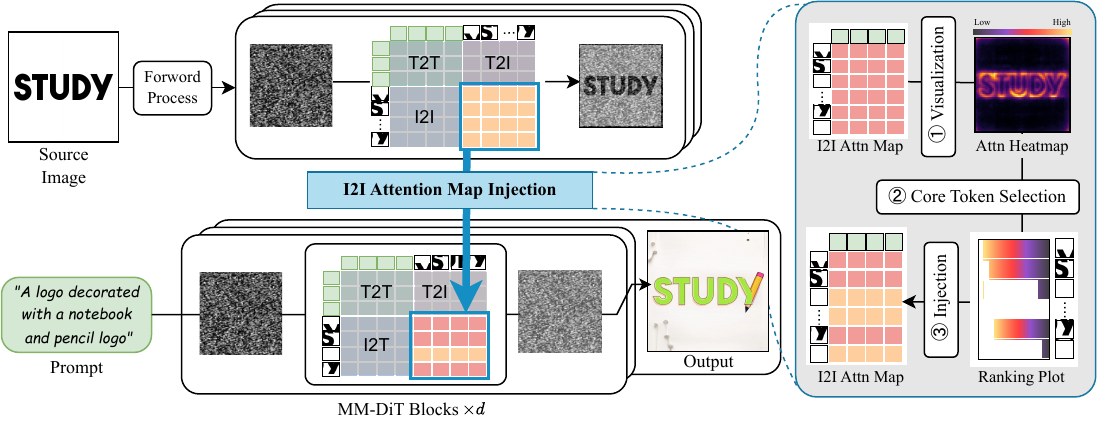}
   \caption{\textbf{Overview of the proposed LogoDiffuser pipeline.} Given an input glyph image $I_s$ and a design prompt $p$, LogoDiffuser selects core tokens from I2I attention within MM-DiT blocks through Core Token selection, and integrates them into the generation process via I2I attention map Injection to ensure that only structure-relevant signals guide the model. Layer-wise Attention Averaging is additionally applied during the injection stage to stabilize structural consistency across layers. These components preserve character shapes faithfully while producing coherent multilingual logo designs.}
   \label{fig:overview}
\end{figure*}
\section{Related Work}
\label{sec:related_work}

\subsection{Text-to-Image Generation}
Early text-to-image diffusion models \cite{rombach2022high, podell2023sdxl, ramesh2022hierarchical, nichol2021glide} primarily adopted U-Net architectures with cross-attention layers to incorporate text conditioning and were trained under DDPM framework \cite{ho2020denoising, dhariwal2021diffusion}. However, convolution-based backbones were limited in global representation and scalability, motivating the transition to transformer-based diffusion models \cite{dosovitskiy2020image, vaswani2017attention}. Recent works such as Diffusion Transformer (DiT) \cite{peebles2023scalable} and PixArt-alpha \cite{chen2023pixartalphafasttrainingdiffusion} demonstrated the scalability and performance of transformer-based designs. Later, Stable Diffusion 3 (SD3) \cite{esser2024scaling} and FLUX \cite{flux2024} introduced MM-DiT, which concatenates text and image tokens into a unified sequence for joint self-attention, enabling coherent and semantically rich image synthesis. Furthermore, conditional control mechanisms, such as ControlNet \cite{zhang2023adding} and IP-Adapter \cite{ye2023ip} allow spatial guidance using edges, depth, or layout. These developments provide a foundation for integrating multimodal attention and structural conditioning.
We use Stable Diffusion 3.5 (SD3.5) \cite{esser2024sd3} as our foundation, whose joint self-attention unifies text–image representations. We further extend token interactions to capture character-level structures essential for multilingual logo generation.

\subsection{Visual Text Generation}

Despite the rapid progress in text-to-image diffusion models \cite{rombach2022high, saharia2022photorealistic, esser2024scaling}, generating legible and visually coherent text within complex scenes remains a significant challenge \cite{saharia2022photorealistic, hu2023tifa}. Existing models struggle to preserve character shapes, especially for non-Latin scripts with intricate stroke structures. 

Recent studies have explored control-based approaches that introduce glyph or layout conditions to guide text rendering. GlyphDraw \cite{ma2023glyphdraw} and GlyphControl \cite{yang2023glyphcontrol} condition the diffusion process on glyph images or positional layouts, improving alignment and legibility. TextDiffuser \cite{chen2023textdiffuser, chen2024textdiffuser} further refines this idea by employing character-level masks and masked-image training to jointly model text synthesis and inpainting. AnyText \cite{tuo2023anytext} extends these frameworks to curved and irregular layouts through multi-branch conditioning. However, these approaches often rely on predefined layouts or fine-tuning, which can limit flexibility and generalization to multilingual languages.
In contrast, we achieve precise character-level control and visually multilingual text rendering without additional modules or training.

\section{Method}
\label{sec:method}

Our goal is to generate multilingual logo images that preserve the structural details of input characters while integrating visual styles harmoniously. Given an input glyph image $I_s$ and a logo design prompt $p$, we aim to synthesize a logo image $I_g$ that maintains the character structure of $I_s$ while reflecting the visual concept described by $p$. To this end, we analyze and control the attention behavior of the multimodal diffusion transformer (MM-DiT) \cite{esser2024scaling}. In Section \ref{subsec:analysis}, we analyze attention maps to identify \textit{core tokens} that respond to character structures. In Section \ref{subsec:core}, we utilize these core tokens for attention map injection to effectively transfer structural information. In Section \ref{subsec:averaging}, we introduce Layer-wise Attention Averaging to maintain consistent structure across layers.

\subsection{Analysis of Image Tokens in MM-DiT}
\label{subsec:analysis}
MM-DiT employs a joint self-attention mechanism that integrates both image and text tokens, with each attention map represented as $qk^T$. In this work, we focus specifically on I2I blocks of MM-DiT, which handle self-attention \cite{dosovitskiy2020image} within the image modality. I2I blocks primarily preserve spatial structure and shape information of the input images, making them critical for maintaining character integrity during logo generation.

\begin{figure}[t!]
    \centering
    \includegraphics[width=\columnwidth]{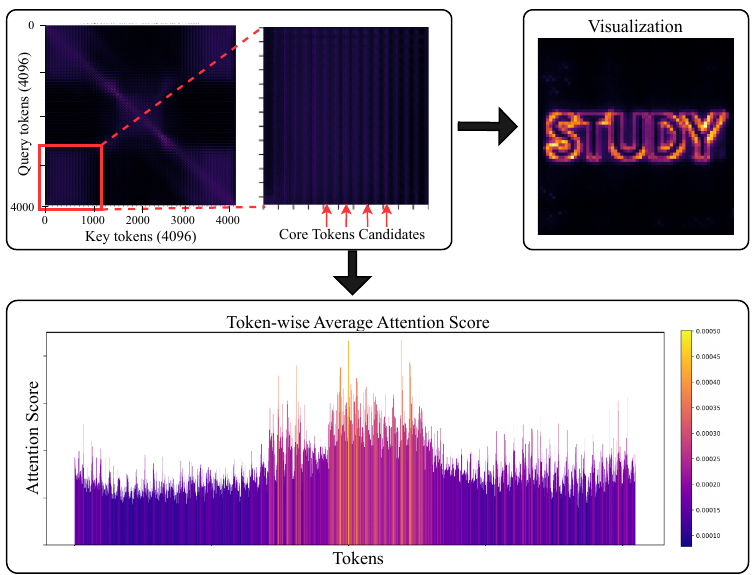}
    \caption {\textbf{Identifying core tokens through token-wise attention analysis.} During glyph image reconstruction, tokens with stronger attention activations concentrate around stroke contours and structural boundaries of the characters. The bottom plot depicts the attention intensity for all tokens, where the highlighted peaks correspond to the most responsive tokens denoted as core token candidates.}
    \label{fig:core_token}
    \vspace{-3mm}
\end{figure}

To analyze the properties of image tokens \cite{dosovitskiy2020image}, we perform image reconstruction using the input glyph images within a Stable Diffusion 3.5 \cite{esser2024sd3} based on MM-DiT. 
During this reconstruction process, we quantitatively and visually examine the generated attention maps. 
As illustrated in Figure \ref{fig:core_token}, we observe that certain image tokens exhibit higher attention responses when reconstructing character shapes. These tokens consistently focus on stroke boundaries and key structural regions of the characters as shown in the upper-right visualization. Furthermore, the lower plot indicates that attention is not uniformly distributed across all tokens, implying that only a subset of tokens contributes significantly to the reconstruction. Based on this observation, we define these highly responsive tokens as \textit{core tokens}, which capture essential glyph structures and faithfully represent the spatial details of the input characters. 
This analysis of I2I blocks provides the foundation for our subsequent method, where core tokens are leveraged to inject structural information into the logo generation process while maintaining stylistic coherence.

\subsection{Core Tokens for Logo Generation}
\label{subsec:core}
In Section \ref{subsec:analysis}, we observe that core tokens respond strongly to the structural features of characters. In this section, we describe how these core tokens can be leveraged to efficiently transfer structural information \cite{vaswani2017attention} during logo generation.

Although the full attention map of MM-DiT \cite{esser2024scaling} captures interactions across all text–image tokens, the information that genuinely reflects glyph structures is concentrated in a subset of tokens with high attention scores. Based on this observation, we selectively inject only the attention maps of the core tokens into the generation process \cite{ho2020denoising}.
As illustrated in Figure \ref{fig:overview}, these core tokens are identified by computing the attention score of all image patch tokens and ranking them in descending order. By selecting the top-k tokens with the highest attention scores, we ensure that the most informative structural signals are preserved.
After identifying these core tokens, we apply attention injection across all attention layers up to a specific timestep.
By focusing solely on core token attention, we filter out non-informative background signals and non-structural noise, propagating only the critical visual relationships associated with character shapes.

\begin{figure}[t!]
    \centering
    \includegraphics[width=\columnwidth]{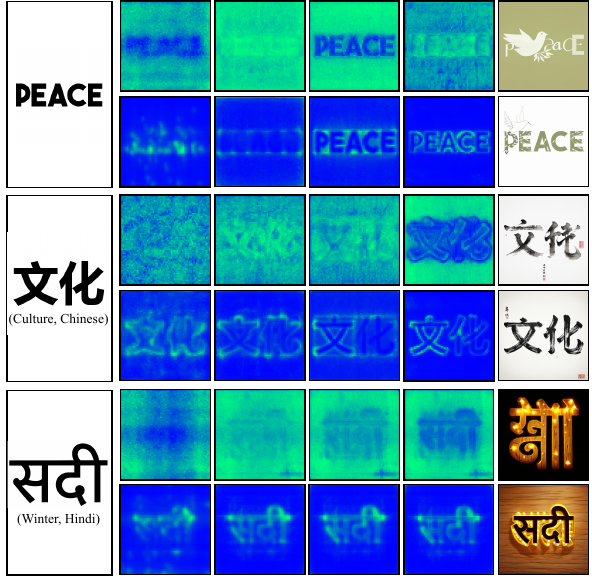}
    \caption{\textbf{Visualization results comparing the full attention maps in the upper row and the core token attention maps in the lower row across three languages.} The core token attention highlights character strokes and boundaries, effectively preserving textual structure while enabling prompt-driven stylization.}
    \label{fig:top_k}
    \vspace{-3mm}
\end{figure}

Figure \ref{fig:top_k} visualizes the difference between the full attention map and the core token attention map. While the full attention is dispersed across areas outside the characters, the core token attention is concentrated on stroke boundaries and key structural regions. This demonstrates that a small subset of tokens with peak attention value tokens is sufficient to preserve character structure information.

Furthermore, as shown in Figure \ref{fig:top_k}, injecting the full attention map results in generated images that inadvertently retain the background from the original character image, which can interfere with producing outputs aligned with the design prompt. In contrast, injecting only the core token attention preserves the original character structure accurately while allowing for stylistic transformations consistent with the prompt.

\subsection{Layer-wise Attention Averaging}
\label{subsec:averaging}

While injecting core token attention effectively preserves character structure, we further analyze an attention shift: not all layers consistently focus on character regions. Specifically, in later layers, the attention of some core tokens tends to shift toward background regions \cite{chefer2021transformer, couairon2023diffedit}. This behavior arises because deeper layers increasingly capture global context, emphasizing visual textures or background elements over structural fidelity.

To address this, we propose a Layer-wise Attention Averaging strategy. Instead of selecting the Top-$k$ core tokens based solely on a single layer’s attention scores, we compute a cumulative average of attention scores across all preceding layers and select the Top-$k$ core tokens based on this averaged map. Formally, the Top-$k$ selection for the $L$-th layer considers not only the attention scores of the current layer but also the accumulated average from layers 1 to $L$. This ensures that local biases in individual layers do not negatively impact the transfer of structural information.

\begin{figure}[t]
  \centering
  \begin{subfigure}{\columnwidth}
    \includegraphics[width=\textwidth]{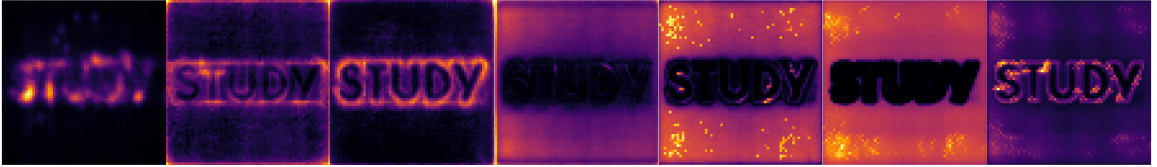}
    \caption{Each Single Layer Attention Map}
    \label{fig:full}
  \end{subfigure}
  \vspace{3mm} 
  \begin{subfigure}{\columnwidth}
    \includegraphics[width=\textwidth]{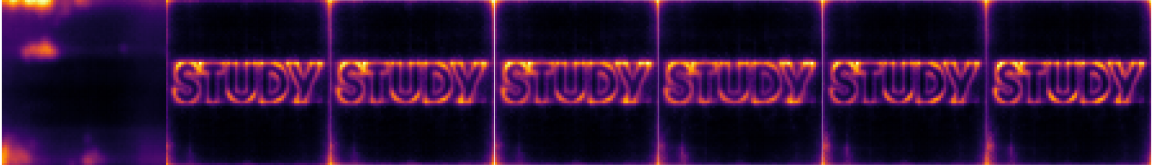}
    \caption{Cumulative Average Attention Map}
    \label{fig:core}
  \end{subfigure}
  \vspace{-3mm}
  \caption{\textbf{Comparison between per-layer attention and cumulative averaged attention.} At step 10, (a) individual layer attention maps attend to different visual regions, while (b) the cumulative average maintains consistent focus on the character structure.}
  \label{fig:full_vs_core}
\end{figure}

This approach enhances the stability of attention selection and preserves structural consistency across layers. As shown in Figure \ref{fig:full}, selecting Top-$k$ tokens based on a single layer can result in high attention on background regions in some layers. In contrast, Figure \ref{fig:core} illustrates that using the averaged Top-$k$ attention ensures consistent focus on character structures across all layers, thereby maintaining the integrity of the input character shapes while supporting stylistic transformations.

\section{Experiments}
\label{sec:experiments}

\subsection{Experimental Setup}
\label{subsec:setup}

\paragraph{Dataset.}
We evaluate our method on a multilingual logo generation task encompassing five languages: English, Chinese, Arabic, Japanese and Korean. This setup allows us to examine how well each model preserves linguistic and stylistic consistency across diverse generative models. For each language, we manually curate 50 representative words. Based on these words, we construct a dataset containing both text prompts and corresponding glyph images, and design prompts in the following format:
\begin{quote}
\textit{``A text \texttt{[word]} logo decorated with \texttt{[style]}.''}
\end{quote}
where \texttt{[word]} and \texttt{[style]} represent the target text in glyph images $I_s$ and the design concept described by the prompt $p$, respectively. Additional details are provided in the supplementary material.

\paragraph{Evaluation Metrics.}
We evaluate the generated results using both quantitative metrics and human evaluation. CLIP score \cite{radford2021learning} measures the semantic alignment between the input prompt and the generated image $I_g$. We further perform an OCR-based quantitative analysis to evaluate the precision of rendered characters. For OCR, we use a large vision-language model Qwen3-VL 32B \cite{yang2025qwen3}. We report accuracy (Acc.), the proportion of samples whose OCR result exactly matches the target word, and F1 score, which considers both precision and recall at the character level.

In addition, we conduct a human evaluation to assess the accuracy and quality of textual rendering in the generated logos. The study is administered via Amazon Mechanical Turk (MTurk).

All methods are evaluated under identical prompt templates and sampling configurations to ensure fair comparison. Each model generates one image per prompt, and no additional sampling or manual selection is performed.

\paragraph{Comparison Methods.}
We compare our method with representative text-to-image diffusion frameworks focusing on controllable or text aware generation. The baselines are grouped into text-rendering-oriented and adapter-based approaches.
We include AnyText \cite{tuo2023anytext} and TextDiffuser-2 \cite{chen2024textdiffuser}, which focus on multilingual and text rendering tasks, respectively. AnyText employs glyph priors and recognition-guided diffusion to preserve character structures, and TextDiffuser-2 integrates segmentation masks and OCR feedback to ensure legibility.
We also use IP-Adapter \cite{ye2023ip} and ControlNet \cite{zhang2023adding}, both applied to SD3. IP-Adapter injects visual features into frozen diffusion models for training-free controllability, while ControlNet provides explicit spatial conditioning through an auxiliary control branch.

\paragraph{Implementation details.}
All experiments of our method are performed using the 12.5\% top-$k$ attention configuration. The sampling is conducted with a guidance scale of 7.5 and 28 diffusion steps. The attention injection is applied to all attention layers within SD3.5 up to the 12th timestep. All experiments are conducted on NVIDIA RTX A5000 GPUs.

\begin{figure*}[t!]
    \centering
    \includegraphics[width=0.95\textwidth]{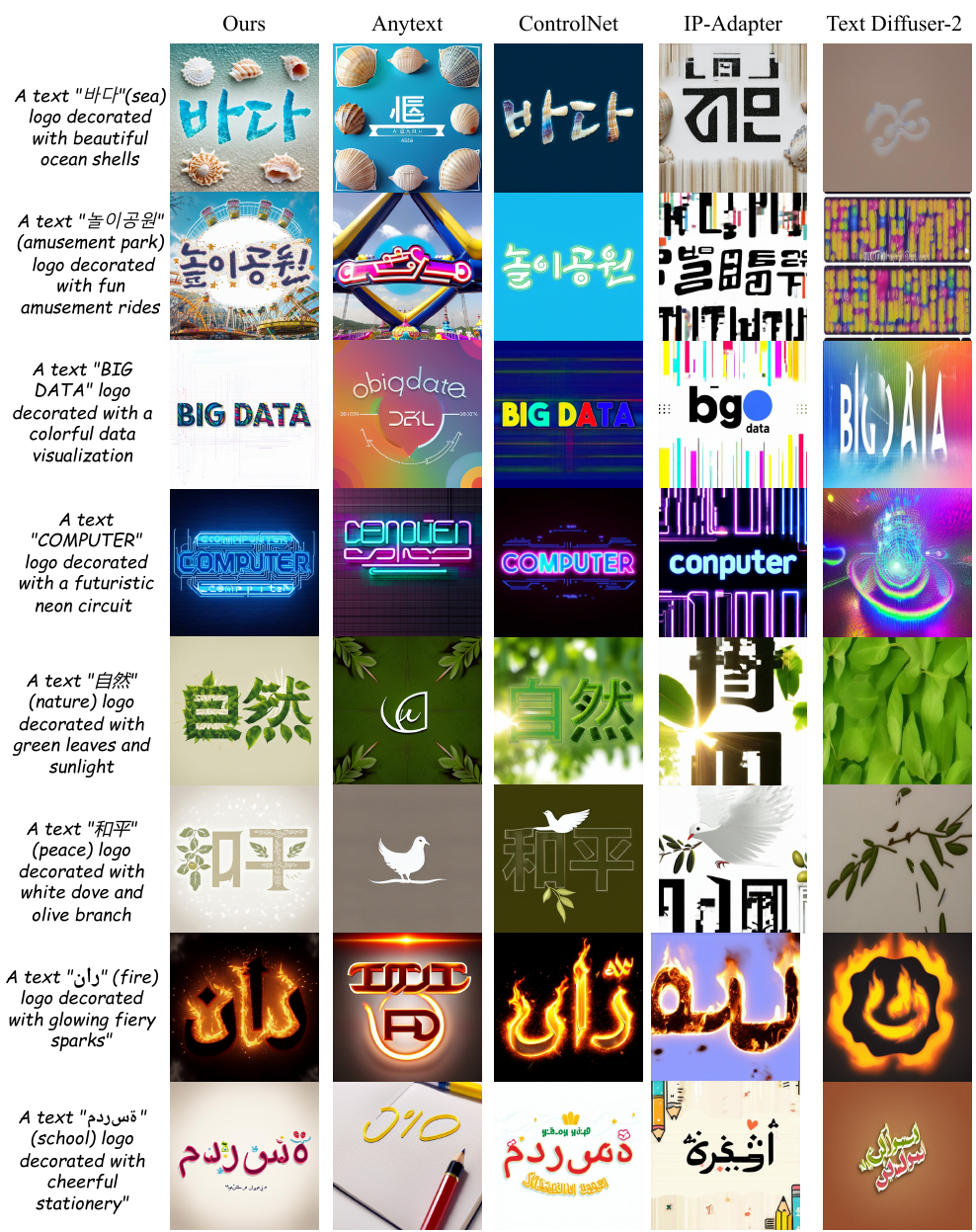}
    \caption {Qualitative results of our method compared to existing approaches.}
    \label{fig:qualitative}
\end{figure*}

\subsection{Qualitative Comparison}
\label{subsec:qualitative}

The generated results illustrate how each model renders textual content across different languages under multilingual logo generation scenarios. Our method produces visually coherent logo designs that accurately reflect both the given text and the intended visual style, as shown in Figure~\ref{fig:qualitative}.

AnyText fails to generate characters reliably and often produces text that is unreadable or inconsistent with the prompt, while its visual style remains only loosely related to the described concept. ControlNet follows stylistic cues well in Arabic prompts but tends to generate incorrect or distorted characters. For Chinese and Korean cases, it reproduces the text more accurately but often neglects the target design concept. IP-Adapter captures general stylistic attributes but does not fully reproduce fine-grained visual details, resulting in partially inconsistent logo styles. TextDiffuser-2 exhibits moderate adherence to the visual concept but fails to consistently reproduce either the design style or the correct characters.

In general, all baselines exhibit stronger performance in generating English text, as most pretrained diffusion and VLM-based architectures are predominantly trained on English-centric datasets \cite{ye2024altdiffusion, nguyen2024multilingual, li2023translation}. However, their ability to render non-Latin scripts such as Korean and Chinese remains limited, often resulting in incomplete or distorted character shapes \cite{wu2019editing, roy2020stefann}. Overall, our method achieves the most balanced performance across languages, preserving text fidelity while accurately following the design intent expressed in the prompt.

\subsection{Quantitative Comparison}
\label{subsec:quantitative}
\begin{table}[t]
    \centering
    \caption{Quantitative comparison of our method compared to existing approaches.}
    \resizebox{\columnwidth}{!}{%
    \begin{tabular}{l|ccccc|cc}
    \toprule
    \multirow{2}{*}{\textbf{Method}} & \multicolumn{5}{c|}{\textbf{CLIP Score}} & \multicolumn{2}{c}{\textbf{Text OCR}} \\
                                     & \textbf{EN} & \textbf{ZH} & \textbf{KO} & \textbf{AR} & \textbf{JA} & \textbf{Acc.} & \textbf{F1} \\
    \midrule
    AnyText            & 24.41 & 24.11 & 22.15 & 23.40 & 24.02 & 0.10 & 0.18 \\
    TextDiffuser-2     & 22.52 & 22.00 & 20.03 & 24.51 & 23.68 & 0.14 & 0.24  \\ \midrule
    IP-Adapter         & 14.45 & 16.85 & 15.20 & 28.11 & 27.61 & 0.46 & 0.63 \\
    ControlNet         & 24.26 & 25.75 & 25.10 & 29.68 & 28.67 & 0.80 & 0.88 \\ \midrule
    \textbf{Ours}      & \textbf{29.43} & \textbf{30.81} & \textbf{27.49} & \textbf{30.31} & \textbf{29.33}  
                       & \textbf{0.80} & \textbf{0.89}  \\
    \bottomrule
    \end{tabular}%
    }
    \label{tab:clip}
\end{table}

    

We report CLIP scores for semantic alignment and OCR-based accuracy and F1 for text legibility. To ensure a fair comparison, all baselines are evaluated on each language individually: EN (English), ZH (Chinese), KO (Korean), AR (Arabic), and JA (Japanese).
As shown in Table~\ref{tab:clip}, our method achieves the best scores, demonstrating stronger prompt–image alignment and clearer character rendering with consistent stylistic coherence.

We examine how the method behaves under different generation conditions by varying the Top-$k$ ratios and diffusion steps for attention injection.
As shown in Table~\ref{tab:top_k_clip}, smaller ratios (12.5\%–25\%) consistently yield stable, high performance, while overly large ratios reduce token selectivity and introduce background noise.
In addition, Tables~\ref{tab:top_k_clip} and~\ref{tab:ocr_en} demonstrate that the 12th diffusion step provides the most stable and reliable results. Even when applying attention injection at other steps, our method maintains comparable performance, indicating its robustness to step selection.

These results show that focusing on a compact set of highly responsive core tokens with the proposed layer-wise attention averaging preserves character structures and stylistic coherence across languages.

\begin{figure}[h!]
    \centering
    \includegraphics[width=0.98\columnwidth]{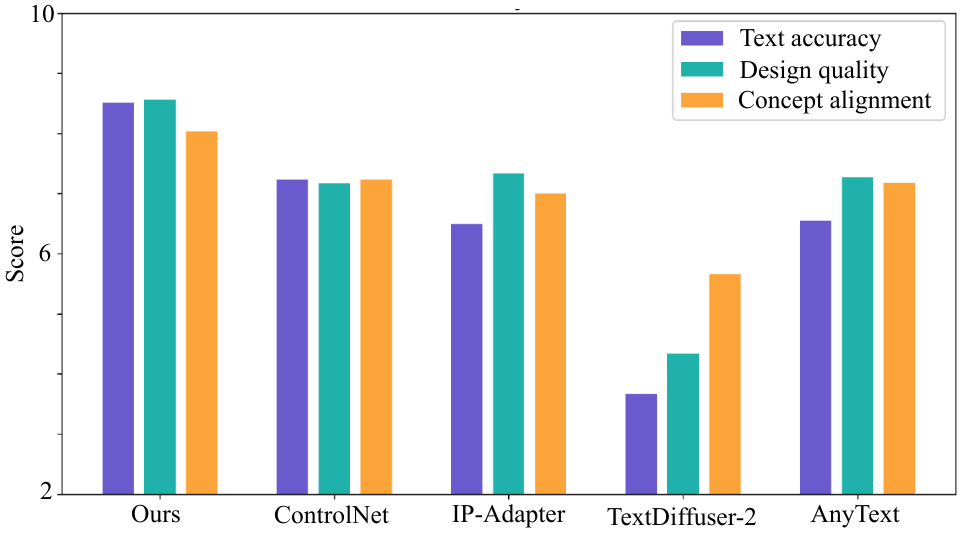}
    \caption {User study comparison between our method and four generation models.}
    \label{fig:user_study}
    \vspace{-3mm}
\end{figure}

\subsection{User Study}
\label{subsec:user_study}
We conducted a user study to evaluate the perceptual quality of logo images generated by different models. The study aimed to assess text accuracy, design quality, and concept alignment—whether the generated logo accurately reflects both the target word and the intended visual style. Each participant was presented with the target word, its corresponding prompt, and the generated logo image, and was asked to rate the image across the three criteria.

Using Amazon Mechanical Turk (MTurk), we collected responses from 100 participants. As summarized in Figure \ref{fig:user_study}, our method received the highest average ratings in all evaluation aspects, demonstrating superior text accuracy (fidelity), design quality and concept alignment compared to existing approaches. Additional details on the user study, including survey questions are provided in the supplementary materials.

\begin{table}[t]
\centering
\renewcommand{\arraystretch}{1.15}
\resizebox{\columnwidth}{!}{%
\begin{tabular}{c|c|ccccc}
\toprule
\multirow{2}{*}{\textbf{Lang.}} & 
\multirow{2}{*}{\textbf{Top-$k$}} & 
\multicolumn{5}{c}{\textbf{Step} } \\
\cmidrule(lr){3-7}
 &  & 8 & 10 & 12 & 15 & 18 \\
\midrule
\multirow{5}{*}{EN}
 & 12.5\%  & 28.9284 & \textbf{29.2844} & \textbf{29.4326} & \textbf{29.2145} & \textbf{29.0422} \\
 & 25\%    & 26.4352 & 28.6722 & 28.3796 & 25.9522 & 25.7202 \\
 & 50\%    & 27.9297 & 28.7461 & 29.4326 & 28.4979 & \textbf{28.3342} \\
 & 75\%    & \textbf{29.1654} & 29.1654 & 28.5436 & 28.5239 & 28.1306 \\
 & 100\%   & 27.7202 & 27.4844 & 26.8419 & 27.0985 & 26.2228 \\

\midrule
\multirow{5}{*}{ZH}
 & 12.5\%   & \textbf{30.3139} & \textbf{30.0836} & \textbf{30.8124} & \textbf{29.7715} & \textbf{29.7838} \\
 & 25\%     & 27.7896 & 29.1185 & 28.2563 & 28.4542 & 28.0768 \\
 & 50\%     & 28.4009 & 28.3941 & 30.8124 & 27.8513 & 27.1479 \\
 & 75\%     & 28.1849 & 28.1849 & 28.0944 & 27.4022 & 25.7951 \\
 & 100\%    & 29.2454 & 29.5766 & 29.2420 & 28.9405 & 25.2127 \\
  
\midrule
\multirow{5}{*}{KO}
 & 12.5\%   & 26.5396 & 26.8592 & 27.4997 & 27.8571 & \textbf{27.9845} \\
 & 25\%     & \textbf{27.6337} & 27.7474 & 27.6675 & 28.1499 & 26.6647 \\
 & 50\%     & 27.3699 & \textbf{28.3121} & \textbf{28.5590} & \textbf{28.3022} & 27.1778 \\
 & 75\%     & 27.0416 & 27.3763 & 26.4044 & 26.9369 & 25.7489 \\
 & 100\%    & 26.5543 & 25.7466 & 25.8106 & 27.2606 & 27.4860 \\

 \midrule
\multirow{5}{*}{AR}
 & 12.5\%   & \textbf{30.3142} & \textbf{30.8255} & \textbf{31.0197} & \textbf{30.3674} & \textbf{29.7229}\\
 & 25\%     & 29.7125 & 29.8122 & 28.4920 & 27.6233 & 27.7396 \\
 & 50\%     & 27.2193 & 27.1589 & 26.7850 & 25.3021 & 25.9305 \\
 & 75\%     & 27.9878 & 26.9814 & 26.0103 & 24.6604 & 24.7511 \\
 & 100\%    & 27.1103 & 26.5452 & 25.3522 & 24.5956 & 23.8708 \\

 \midrule
\multirow{5}{*}{JA}
 & 12.5\%   & 29.3348 & 28.5256 & 28.9844 & \textbf{29.0127} & \textbf{30.1507} \\
 & 25\%     & \textbf{30.3474} & \textbf{29.1085} & \textbf{29.6518} & 28.9441 & 27.8603 \\
 & 50\%     & 29.3542 & 28.1044 & 27.7257 & 26.4961 & 26.5675 \\
 & 75\%     & 28.4980 & 28.0463 & 27.6607 & 26.6633 & 25.6138 \\
 & 100\%    & 27.7093 & 28.0829 & 27.6330 & 25.7725 & 25.3619 \\
\bottomrule
\end{tabular}%
}
\caption{Quantitative results under different Top-$k$ ratios across diffusion steps and languages.}
\label{tab:top_k_clip}
\end{table}

\begin{table}[h!]
\centering
\footnotesize
\setlength{\tabcolsep}{8pt} 
\renewcommand{\arraystretch}{0.8}       
\begin{tabular}{c|ccccc}
\toprule
\multirow{2}{*}{\textbf{Metric}} &
\multicolumn{5}{c}{\textbf{Step}} \\
\cmidrule(lr){2-6}
 & 8 & 10 & 12 & 15 & 18 \\
\midrule
 Acc. & 0.72 & 0.72 & 0.80 & 0.82 & 0.88 \\
 F1   & 0.83 & 0.83 & 0.89 & 0.90 & 0.93 \\
\bottomrule
\end{tabular}
\caption{Quantitative results of English OCR accuracy and F1 across diffusion steps.}
\label{tab:ocr_en}
\end{table}

\subsection{Applications}
\label{subsec:applications}

\begin{figure}[h!]
    \centering
    \includegraphics[width=\columnwidth]{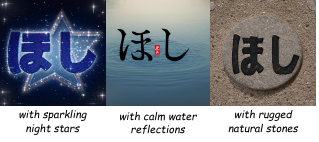}
    \caption {Robustness to font variation in logo generation. The \texttt{[word]} used is \begin{CJK}{UTF8}{min}ほし\end{CJK}(star), and the \texttt{[style]} descriptions are shown below each generated image.}
    \label{fig:font}
    \vspace{-3mm}
\end{figure}

\paragraph{Adapting to Different Fonts and Design Styles.}  
We evaluate robustness to font variation using the Japanese word \begin{CJK}{UTF8}{min}ほし\end{CJK} (star) under various stylistic conditions.  
As shown in Figure~\ref{fig:font}, our model consistently preserves structure and readability across different typefaces, effectively adapting to diverse design styles.

\begin{figure}[h!]
    \centering
    \includegraphics[width=\columnwidth]{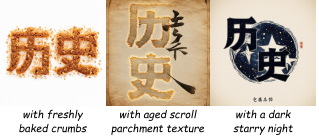}
    \caption {Robustness to text position variation. The \texttt{[word]} used is \begin{CJK}{UTF8}{gbsn}历史\end{CJK}(history), and the \texttt{[style]} descriptions are shown below each generated image.}
    \label{fig:location}
    \vspace{-3mm}
\end{figure}

\paragraph{Maintaining Consistency Across Text Positions.}  
To assess spatial robustness, we vary the placement of the Chinese word \begin{CJK}{UTF8}{gbsn}历史\end{CJK} (history) horizontally, vertically, and diagonally.  
As illustrated in Figure~\ref{fig:location}, our method maintains semantic coherence and stylistic harmony regardless of spatial placement.

\begin{figure}[h!]
    \centering
    \includegraphics[width=\columnwidth]{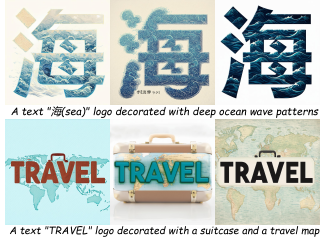}
    \caption {Diverse generation results from identical prompts and target words. Prompts are shown below each generated images.}
    \label{fig:same_word_same_prompt}
    \vspace{-3mm}
\end{figure}

\paragraph{Diversity in Multilingual Logo Generation.}  
We examine the creative diversity of our model by generating multiple logo images from identical prompts and target words using different random seeds.  
As shown in Figure~\ref{fig:same_word_same_prompt}, our method produces visually distinct yet semantically consistent outputs, demonstrating strong expressive diversity while maintaining conceptual integrity.

\section{Conclusion}
\label{sec:conclusion}

In this work, we introduce LogoDiffuser, a training-free method for generating multilingual logo designs that integrate textual and visual elements within MM-DiT. By treating characters as image inputs instead of textual prompts, our method enables accurate preservation of character structures across different languages. By analyzing the attention features of MM-DiT, we identify core tokens that strongly respond to character regions and play an important role in maintaining structural details. Based on this observation, LogoDiffuser injects the attention of these core tokens to combine clear character shapes with creative visual styles during generation. In addition, we introduce the Layer-wise Attention Averaging strategy that stabilizes attention across layers and ensures consistent structure. Through extensive experiments and user studies, our method demonstrates significant improvements in visually faithful designs with accurate text rendering. These findings highlight the potential of core token–based attention control to enhance text fidelity in multilingual visual text synthesis suggest promising directions for future research in text-to-Image generation.

{
    \small
    \bibliographystyle{ieeenat_fullname}
    \bibliography{main}
}

\clearpage
\setcounter{page}{1}
\maketitlesupplementary


\section{Dataset}
\label{sec:dataset}
\begin{figure}[t!]
    \centering
    \includegraphics[width=1.1\columnwidth]{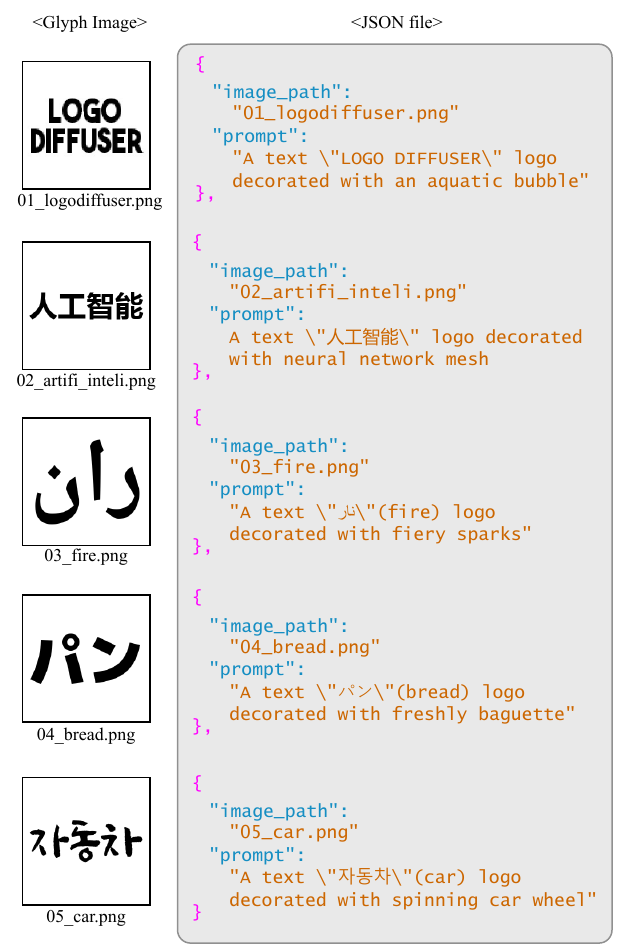}
    \caption {Examples from our dataset, showing glyph images and their corresponding JSON-formatted prompts.}
    \label{fig:sup_dataset}
\end{figure}

Figure \ref{fig:sup_dataset} shows a sample of our dataset. For each language, we collected 50 representative words and rendered them as glyph images $I_s$ to serve as input for logo generation. The corresponding text prompts $p$ are constructed in the format:
\begin{quote}
\textit{``A text \texttt{[word]} logo decorated with \texttt{[style]}.''}
\end{quote}
We built five separate datasets for English, Chinese, Japanese, Arabic, and Korean. A wide variety of logo styles were included to ensure diverse visual appearances, covering different artistic concepts, typographic influences, and decorative patterns. This dataset enables evaluation of both glyph preservation and style adherence across multiple languages and design scenarios.

\section{Additional Results}
\label{sec:add_results}
\begin{figure*}[t!]
    \centering
    \includegraphics[width=0.9\textwidth]{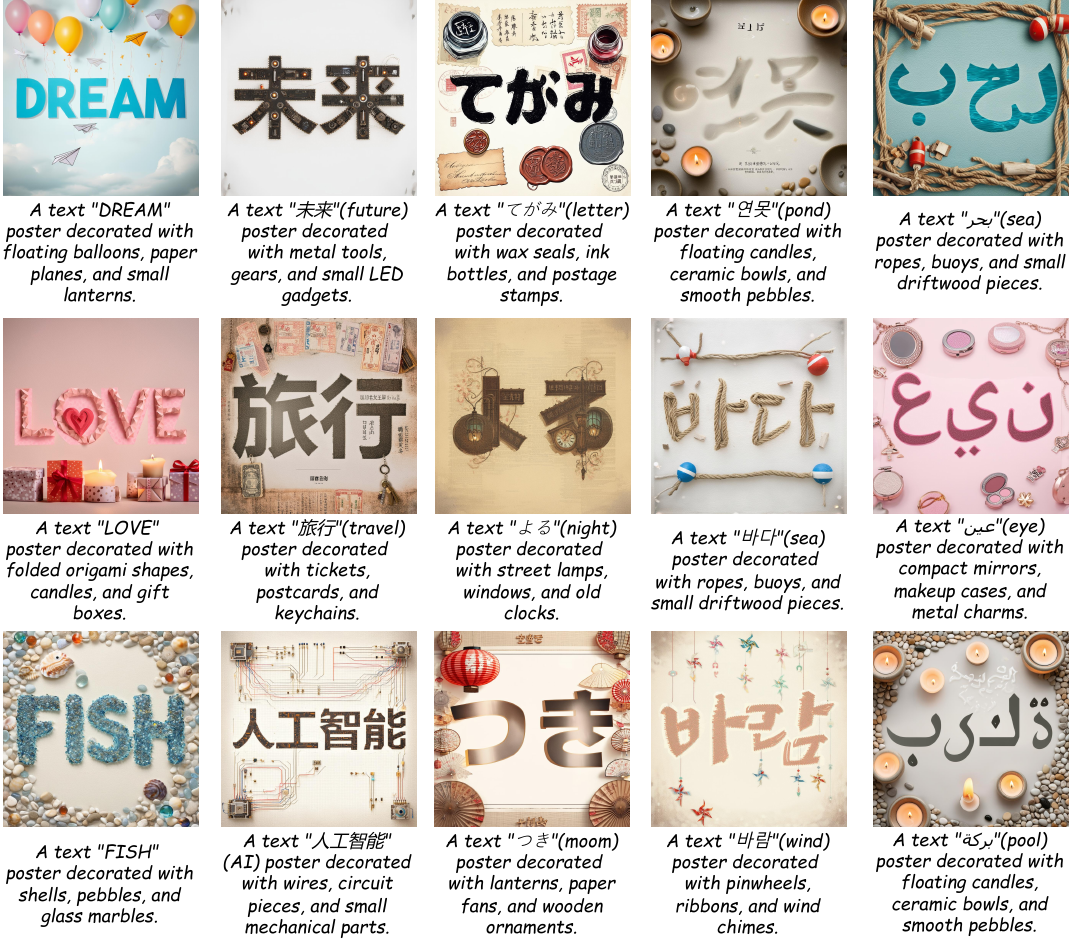}
    \caption {Additional results of LogoDiffuser.}
    \label{fig:supp_other_prompts}
\end{figure*}

We present additional result in Figure \ref{fig:supp_other_prompts}, demonstrating the applicability of our method beyond logo generation. In addition to the prompts used in our dataset, we also evaluate the model with alternative prompt formulations to examine its generalization to broader visual design tasks. Specifically, we assess its ability to synthesize poster-style images that require coherent integration of text, layout structure, and global visual composition. Our method generates clear and faithful text while maintaining the intended visual appearance. The generated posters preserve the structure of the input characters and follow the described style without introducing unintended artifacts. These results indicate that the proposed approach generalizes well to broader design tasks beyond logo generation.

\section{User Study Details and Inter-Rater Agreement}
\label{sec:supp_user_study}
\begin{figure*}[t!]
    \centering
    \includegraphics[width=0.5\textwidth]{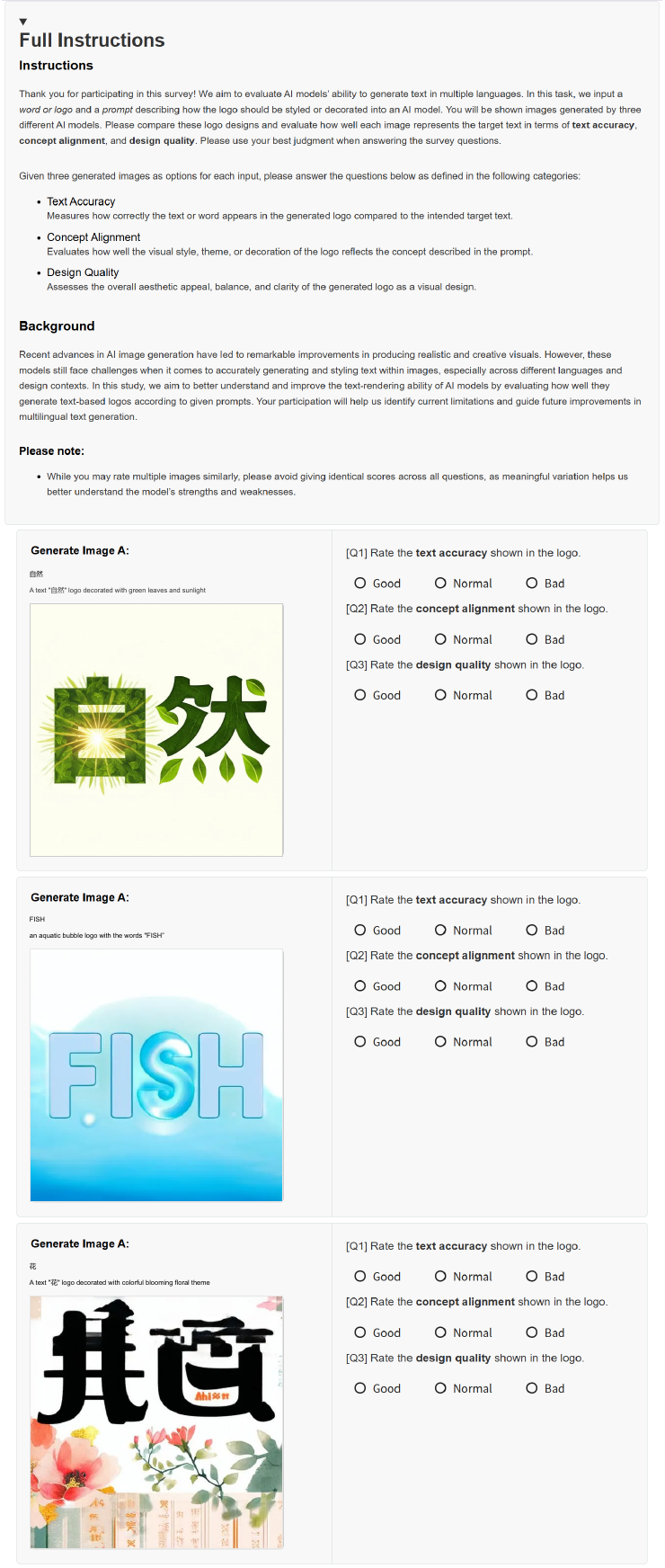}
    \caption {MTurk evaluation form. Participants rated three randomly ordered logo images for the same prompt according to text accuracy, concept alignment, and design quality.}
    \label{fig:supp_user_study_example}
\end{figure*}

We conducted a human evaluation using Amazon Mechanical Turk (MTurk) to compare our method against existing baselines. 
Each HIT presented three logo images for the same target text and prompt: one generated by our method and two randomly selected baseline methods. An example of the MTurk evaluation interface is shown in Figure \ref{fig:supp_user_study_example}.
The order of the three images was fully randomized for every HIT to prevent positional bias.
For each image, participants rated three criteria: (1) text accuracy, (2) concept alignment, and (3) design quality.

To assess annotation consistency, we computed pairwise Cohen’s $\kappa$ across all participants pairs.
The average agreement was low (text accuracy: $\kappa=0.018$, concept alignment: $\kappa=0.028$, design quality: $\kappa=0.013$),
indicating that annotators made independent and non-coordinated judgments.
Given the inherently subjective and perceptual nature of the task, low agreement is expected and even desirable,
as it reflects genuine diversity in human perception rather than a dominant or biased labeling pattern.

\end{document}